%% file: GARO.tex
\pdfoutput=1

\documentclass[letterpaper, 10pt, conference]{ieeeconf}
 
\IEEEoverridecommandlockouts                              

\usepackage{multicol}
\usepackage{multirow}
\usepackage{algpseudocode}
\usepackage{color}
\usepackage{graphicx}
\usepackage{bm}
\usepackage{amsmath}
\usepackage{amssymb}
\usepackage{tabularx}
\usepackage{subcaption}
\usepackage{url}
\usepackage{cite}
\usepackage{prettyref}
\usepackage{booktabs}
\usepackage{siunitx}
\usepackage{comment}
\usepackage{booktabs}
\usepackage{hyperref}
\usepackage{tabularx, booktabs}
\usepackage[capitalise]{cleveref}

\usepackage{makecell} 
\usepackage[table]{xcolor} 
\usepackage{pifont} 
\usepackage{tabularx} 
\definecolor{Gray}{gray}{0.9}

\usepackage[all]{hypcap}
\usepackage{comment}

\usepackage[ruled]{algorithm2e} 

\SetAlFnt{\small}
\SetAlCapFnt{\small}
\SetAlCapNameFnt{\small}
\SetAlCapHSkip{0pt}

\newrefformat{fig}{Figure~\ref{#1}}
\newrefformat{sec}{Section~\ref{#1}}
\newrefformat{tab}{Table~\ref{#1}}
\newrefformat{alg}{Algorithm~\ref{#1}}
\newrefformat{eqn}{Equation~\ref{#1}}

\colorlet{best}{red!25}
\colorlet{second}{orange!25}
\colorlet{third}{yellow!25}

\hypersetup{
    colorlinks,
    linkcolor={blue!80!black},
    citecolor={blue!80!black},
    urlcolor={blue!100!black}
}

\makeatletter
\newcommand{\printfnsymbol}[1]{%
  \textsuperscript{\@fnsymbol{#1}}%
}
\makeatother

\title{\LARGE \bf GARO: Geometry-Aware Redundancy Optimization for Real-Time and High-Fidelity Dynamic Gaussian Splatting
\thanks{\copyright\ 2026 IEEE. Personal use of this material is permitted. Permission from IEEE must be obtained for all other uses, in any current or future media, including reprinting/republishing this material for advertising or promotional purposes, creating new collective works, for resale or redistribution to servers or lists, or reuse of any copyrighted component of this work in other works. Accepted for publication in Proceedings of the 2026 IEEE International Conference on Robotics and Automation (ICRA 2026).}}

\author{
    Huiwen Xue$^{1}$ \and
    Kaixing Zhao$^{1,\ast}$ \and
    Zuheng Ming$^{2,3}$ \and
    Tingcheng Li$^{4}$\\
    \and
    $^{1}$School of Software, Northwestern Polytechnical University
    \and
    $^{2}$L2TI, Université Sorbonne Paris Nord 
    \and 
    $^{3}$EmboMind Research
    \and
    $^{4}$School of Electronic Information and Engineering, Suzhou University of Science and Technology
    \thanks{$^{*}$~The corresponding authors}
}
\begin{document}
\maketitle

\begin{abstract}
    
Novel view synthesis is a key task for dynamic scene reconstruction, where high rendering speed is essential for applications such as virtual reality. Existing deformable Gaussian Splatting methods achieve high-fidelity dynamic scene modeling, but still face limitations in memory usage and rendering efficiency due to the large number of redundant Gaussians. To address these challenges, we propose Geometry-Aware Redundancy Optimization (GARO), a unified redundancy measurement framework in the adaptive density control stage of the traditional dynamic scene reconstruction pipeline. This framework first selects low-gradient candidates using an optimization activity assessment strategy, and then evaluates geometric complexity through low curvature analysis to further filter and prune redundant points, resulting in a compact and expressive Gaussian representation. Extensive experiments on synthetic and real-world datasets demonstrate that GARO achieves robust trade-offs between quality and speed, with PSNR remaining stable and rendering speed improved by $2\times$, validating the efficiency and effectiveness of GARO.

\end{abstract}

\input{tex/Intro}

\input{tex/Related_work}

\input{tex/method}
\input{tex/Experiments}
\input{tex/Conclusion}
{\small
\bibliographystyle{IEEEtran}
\addtolength{\textheight}{-0.5cm}
\bibliography{references}
}

\end{document}

%% file: tex/Intro.tex
\section{Introduction}

Novel view synthesis plays a critical role in applications such as autonomous driving \cite{LidaRF}, virtual reality \cite{FastScene}, and robotics \cite{Palm}, where high frames-per-second (FPS) rendering is increasingly required. The field has evolved from geometric reconstruction to neural implicit and explicit representations. Neural Radiance Fields (NeRF) \cite{Nerf} significantly improved detail, but remain limited by slow rendering and high computational cost. 3D Gaussian Splatting (3DGS) \cite{3dgs} introduces an explicit representation with 3D Gaussians and a differentiable rendering pipeline, greatly improving rendering efficiency. Recent research \cite{Dynamic,Mip-Splatting,IRGS} has further enhanced 3DGS rendering quality by addressing diverse challenges.
\begin{figure}[!htb]
\centering 
\includegraphics[width=0.85\linewidth]{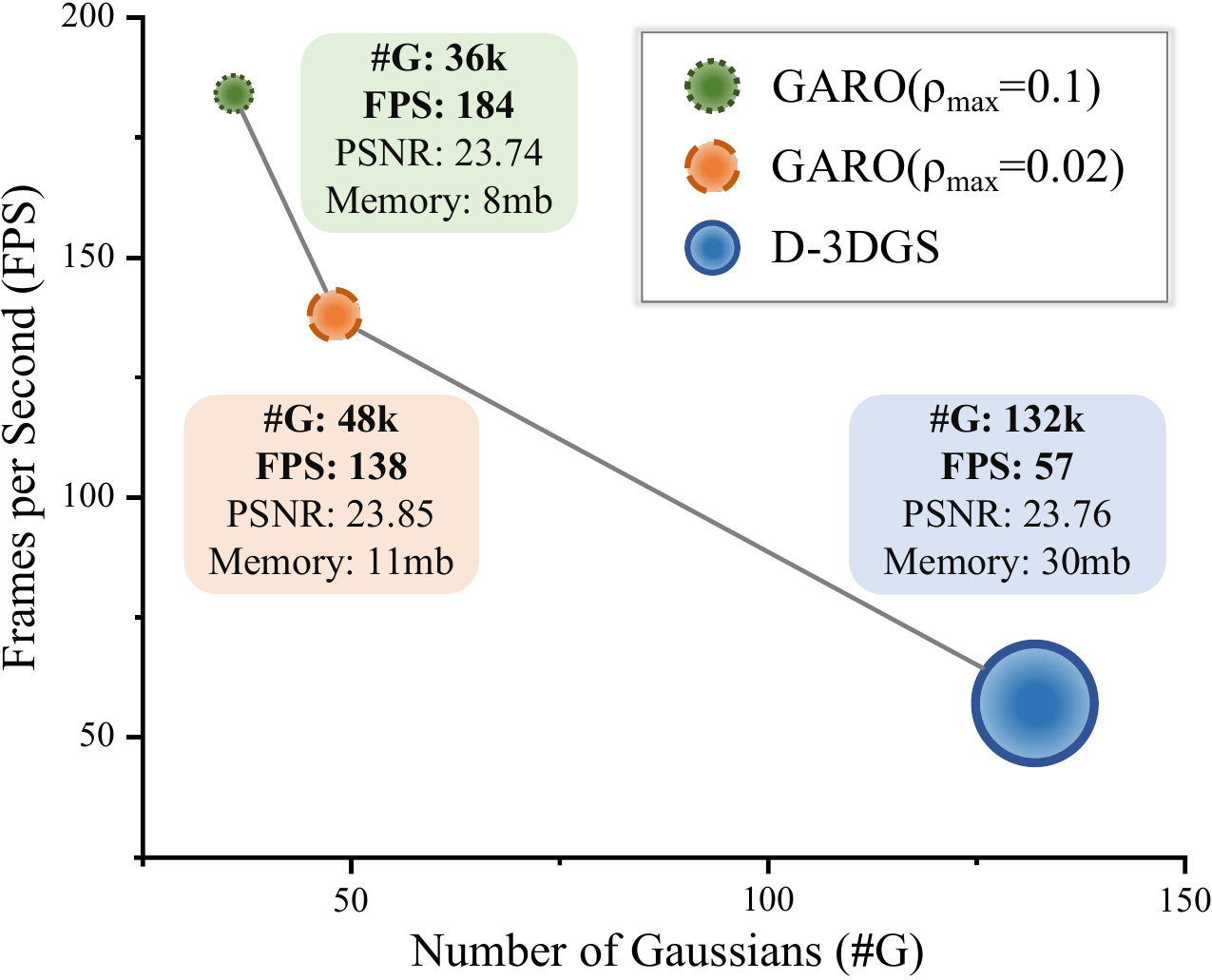}
  \caption{\textbf{Relationship between FPS and number of Gaussians.} By reducing redundant Gaussians across a maximum pruning ratio $\rho_{max}$, GARO achieves over $2\times$ speedup and reduces memory consumption indicated by bubble radius.}
  \label{fig1}
\end{figure}
\begin{figure*}[htbp]
\centering 
\includegraphics[width=0.95\textwidth]{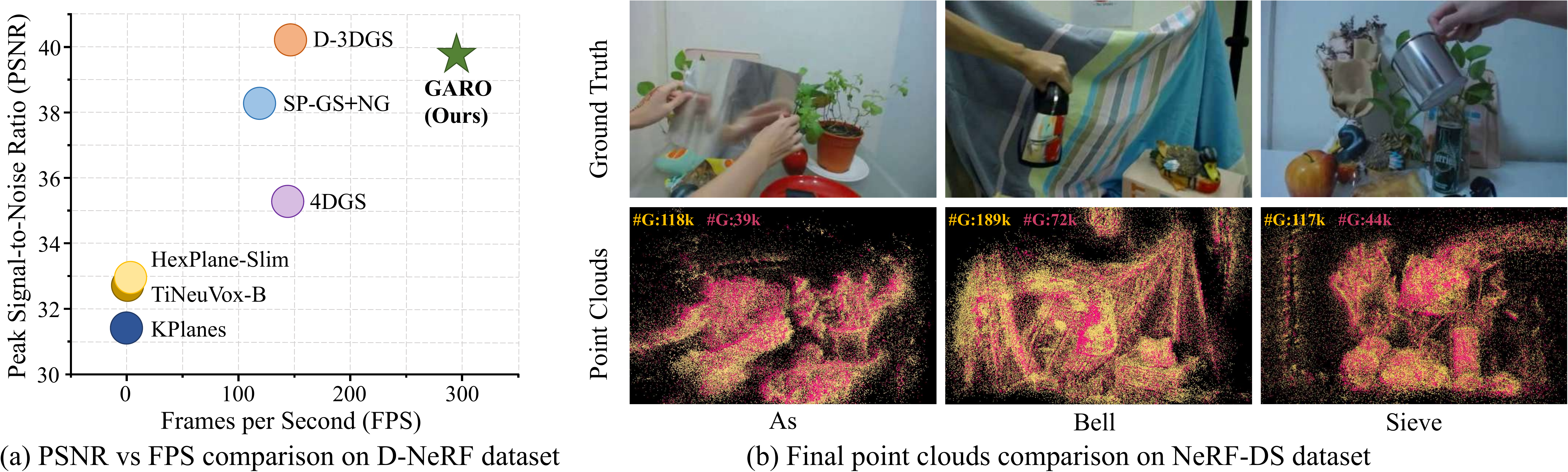}
    \caption{(a) Our GARO achieves the best FPS while maintaining high rendering quality. (b) Yellow points represent the results of D-3DGS, while red points indicate the output of our GARO. \#G denotes the number of Gaussians. The point clouds generated by GARO are mainly concentrated in high-curvature regions of the objects, significantly reducing redundant points in flat areas.} 
\label{teaser}
\end{figure*}

Although 3DGS enables real-time, high-fidelity rendering, high memory usage and slow inference still limit its efficiency in complex dynamic scenes. Recent works address these issues through four main directions to improve efficiency. Codebook-based methods \cite{LightGaussian,CompGS,Compressed,Reducing} compress Gaussian parameters for efficient storage and transmission. Pruning approaches \cite{HAC,PUP,Compact,LP-3DGS,Radsplat} directly remove redundant Gaussians to reduce memory and computation. Densification and spatial optimization \cite{Mini-splatting,Taming} adaptively adjust Gaussian distributions to balance detail and efficiency. Structural and generative methods \cite{Scaffold-GS,DiffGS} enhance flexibility and expressiveness through anchor-based or generative modeling.

However, these existing approaches focus on static scenes with fixed geometry and attributes, lacking temporal modeling for dynamic scenarios. Recent works address these challenges by modeling time-varying parameters \cite{Compact_dy}, introducing probabilistic pruning and masked rasterization \cite{MaskGaussian}, filtering static regions with sparse anchors \cite{EDGS}, or clustering Gaussians into superpoints for efficient manipulation \cite{SP-GS}.

Among these existing approaches, pruning-based optimization is a key strategy for improving rendering efficiency in 3DGS, as it directly reduces redundant Gaussians and lowers storage and computational costs. Our work adopts intelligent pruning to further enhance dynamic scene efficiency. Redundant Gaussians are a major cause of storage and rendering inefficiency. Since each Gaussian and its attributes require independent storage and optimization, this leads to high memory and computation overhead, significantly reducing rendering speed. As shown in Fig.~\ref{fig1}, increasing the number of Gaussians clearly decreases rendering efficiency. For the same scene, Deformable 3D Gaussian Splatting (D-3DGS) uses 132k Gaussians at 57 FPS, while our method reduces this to 36k, achieving 184 FPS, with comparable quality. These results confirm that Gaussian redundancy is a core bottleneck for 3DGS efficiency.

Moreover, existing pruning methods typically rely on heuristics like parameter variation or importance scores, often neglecting the underlying optimization activity and geometric structure, thus limiting efficiency gains.
Unlike these methods, we analyze redundant Gaussians from a novel perspective. Dynamic scene reconstruction suffers from the accumulation of Gaussians with both low optimization activity and low geometric complexity. These primitives, which undergo minimal parameter updates during training and are located in locally flat surface regions, may satisfy traditional pruning requirements but contribute little to final rendering quality while consuming substantial computational resources.
However, the pruning mechanism in 3DGS primarily relies on coarse-grained metrics such as opacity thresholds and spatial size, lacking sophisticated geometric analysis capabilities.
While recent work like RT-GuIDE \cite{RTGuIDE} utilizes gradient-based metrics to quantify information gain for exploration, the potential of combining such optimization activity with geometric structure for pruning in dynamic scenes remains under-explored.

To address these limitations, we introduce a unified redundancy measurement framework that assesses redundancy from two complementary perspectives. The Optimization Activity Assessment, based on gradient magnitude, reflects the parameter update extent during training, while the Geometric Complexity Assessment, based on curvature, distinguishes redundant Gaussians in flat regions from essential ones in curved areas. 
By jointly applying these two assessment strategies, GARO achieves a superior balance between quality and efficiency, as shown in Fig.~\ref{teaser}(a). 
Specifically, our adaptive pruning mechanism prunes Gaussians based on the real-time gradients and curvature of each Gaussian during training, enabling it to dynamically respond to the evolving distribution of these properties. This ensures efficient resource allocation and robust scene representation across diverse scenarios. As shown in Fig.~\ref{teaser}(b), the final point clouds comparison with D-3DGS clearly demonstrates that GARO effectively prunes redundant points while preserving key Gaussians, validating the effectiveness of our geometry-aware redundancy optimization strategy. 
To summarize, our contributions are as follows:
\begin{itemize}
\item We propose a unified redundancy measurement framework that effectively combines optimization activity and geometric complexity for dynamic 3D Gaussian redundancy identification, enabling more comprehensive and effective pruning.
\item We design an adaptive pruning strategy that efficiently reduces the number of redundant Gaussians and memory consumption while preserving rendering quality and improving efficiency in dynamic scenes.
\item  Our GARO achieves more than a $2\times$ speedup in rendering, while maintaining comparable or even superior reconstruction quality.
\end{itemize}

\label{sec:intro}

%% file: tex/Related_work.tex
\section{Related Work}
\subsection{Novel View Synthesis}
Early methods based on Structure from Motion (SfM) \cite{sfm} and Multi-View Stereo (MVS) \cite{mvs} relied on explicit geometry but struggled with complex lighting and materials. Neural Radiance Fields (NeRF) \cite{Nerf} introduced neural implicit representations using MLPs for volumetric density and radiance, achieving high-quality rendering. Extensions such as D-NeRF \cite{D-NeRF}, NeRF-DS \cite{NeRF-DS}, Nerfies \cite{Nerfies}, HyperNeRF \cite{HyperNeRF}, Ti-DNeRF \cite{ti-nerf}, and TiNeuVox \cite{TiNeuVox-B} improved dynamic modeling and efficiency. Methods like Tensor4D \cite{Tensor4D}, K-Planes \cite{K-Planes}, and HexPlane \cite{HexPlane} further balanced quality and speed, but real-time performance remains challenging.
3D Gaussian Splatting (3DGS) \cite{3dgs} achieved a breakthrough in real-time rendering by representing scenes with explicit 3D Gaussians. Extensions to dynamic scenes include time-varying modeling \cite{Spacetime}, deformation-based modeling \cite{D-3DGS}, and 4D Gaussian modeling \cite{4DGS_Yang,4DGS_wu}. Further improvements in rendering quality have been made, with SC-GS \cite{SC-GS} leveraging sparse control points for spatiotemporal consistency and editability, FDS-GS \cite{FDS-GS} introducing explicit density and scale constraints for enhanced detail, and Mip-Splatting \cite{Mip-Splatting} applying frequency-aware constraints and filtering to suppress high-frequency artifacts. Despite these advances, challenges remain in rendering efficiency, highlighting the need for more effective Gaussian optimization mechanisms.

\subsection{Efficient Gaussian Splatting}

To reduce the memory overhead of 3DGS, various methods have been proposed. For parameter compression, LightGaussian \cite{LightGaussian} and CompGS \cite{CompGS} use importance scores, vector quantization, and clustering to compress Gaussian parameters, while Compressed 3D Gaussian \cite{Compressed} and Reducing \cite{Reducing} further optimize with sensitivity-based clustering and adaptive codebooks. For point count and distribution, HAC \cite{HAC}, PUP 3D-GS \cite{PUP}, Compact3DGS \cite{Compact}, LP-3DGS \cite{LP-3DGS}, and RadSplat \cite{Radsplat} employ context modeling, adaptive masking, and learnable masks to efficiently prune and maintain scene quality. Mini-Splatting \cite{Mini-splatting} and Taming 3DGS \cite{Taming} improve spatial densification and model size through guided strategies. Scaffold-GS \cite{Scaffold-GS} and DiffGS \cite{DiffGS} enhance Gaussian organization and generation via anchor points and diffusion models. While these approaches accelerate rendering, most are limited in dynamic scenarios with temporal changes. Recent works address dynamic efficiency through temporal modeling \cite{Compact_dy}, probabilistic pruning \cite{MaskGaussian}, sparse anchors \cite{EDGS}, and clustering \cite{SP-GS}. Unlike prior pruning methods, our GARO identifies redundant Gaussians from both geometric and optimization perspectives, enabling effective pruning and efficient dynamic reconstruction.

%% file: tex/method.tex
\section{Preliminaries}

\subsection{3D Gaussian Splatting}
3D Gaussian Splatting (3DGS) \cite{3dgs} uses explicit point clouds to model scene geometry and appearance. Each Gaussian is characterized by a center position $\boldsymbol{\mu} \in \mathbb{R}^3$ and a covariance matrix $\Sigma \in \mathbb{R}^{3 \times 3}$:
\begin{equation}
    G(\mathbf{x}) = e^{-\frac{1}{2}(\mathbf{x} - \boldsymbol{\mu})^T \Sigma^{-1} (\mathbf{x}-\boldsymbol{\mu})}
.
\end{equation}
To ensure the positive semi-definiteness of the covariance matrix, 3DGS parameterizes $\Sigma$ through a rotation matrix $R$ and a scaling matrix $S$:
\begin{equation}
\Sigma =R S S^T R^T 
.
\end{equation}
3D Gaussians are rendered as elliptical shapes on the image plane, with their colors and opacities computed by blending their projections according to depth:
\begin{equation}
C(u,v) =  \sum_{i=1}^{N} c_i \cdot \alpha_i \cdot \prod_{j=1}^{i-1}(1-\alpha_j) \cdot G_i(u,v) 
,
\end{equation}
where $G_i(u,v)$ represents the 2D projection of Gaussian $i$ onto the image plane, $c_i$ denotes its color, and $\alpha_i$ is the opacity. The training process of 3DGS is driven by a photometric loss function that combines L1 distance and structural similarity: 
\begin{equation}
\mathcal{L}_{RGB} = (1-\lambda_{DSSIM})\mathcal{L}_1 + \lambda_{DSSIM}(1-SSIM)
\label{eq:L_RGB}
,
\end{equation}
where $\mathcal{L}_1$ measures the L1 distance between rendered and ground truth images, while $SSIM$ evaluates structural similarity.

Additionally, 3DGS incorporates an adaptive density control mechanism to maintain an optimal Gaussian distribution. The densification process clones Gaussians with small scales in high-gradient regions, while splitting oversized Gaussians into multiple smaller ones. Conversely, the pruning mechanism removes Gaussians with low opacity or excessive screen projections to maintain rendering efficiency.

\subsection{Deformable 3D Gaussian Splatting}

For dynamic scene modeling, Deformable 3D Gaussian Splatting (D-3DGS) \cite{D-3DGS} extends static 3DGS by incorporating a time-conditional deformation field. The framework employs a deformation MLP network $\mathcal{F}_{\theta}$, which predicts position, rotation, and scaling offsets based on canonical Gaussian properties and temporal information: 
\begin{equation}
(\delta\mathbf{x}, \delta\mathbf{r}, \delta\mathbf{s}) = \mathcal{F}_{\theta}(\gamma(\mathbf{x},t))
.
\end{equation}
Here, $\mathbf{x}$ denotes the canonical Gaussian position, $t$ represents the normalized frame index that drives deformation, and $\gamma$ indicates positional encoding. The network outputs enable dynamic transformation of static parameters across temporal sequences.

Despite achieving high-quality dynamic scene reconstruction, D-3DGS still faces significant performance bottlenecks in practice due to the rapid growth of Gaussians during training. The resulting redundancy greatly increases computational complexity and memory consumption, hindering real-time rendering and limiting deployment on resource-constrained devices. Therefore, more intelligent and efficient optimization strategies are crucial for improving rendering performance in dynamic reconstruction scenarios.

\section{Method}
To address the performance bottlenecks caused by redundant Gaussians in dynamic scene reconstruction, we propose Geometry-Aware Redundancy Optimization (GARO), a novel approach that leverages geometric structure analysis for efficient redundancy identification while preserving reconstruction quality. 

\begin{figure*}[t]
\centering
\framebox{\includegraphics[width=0.95\textwidth]{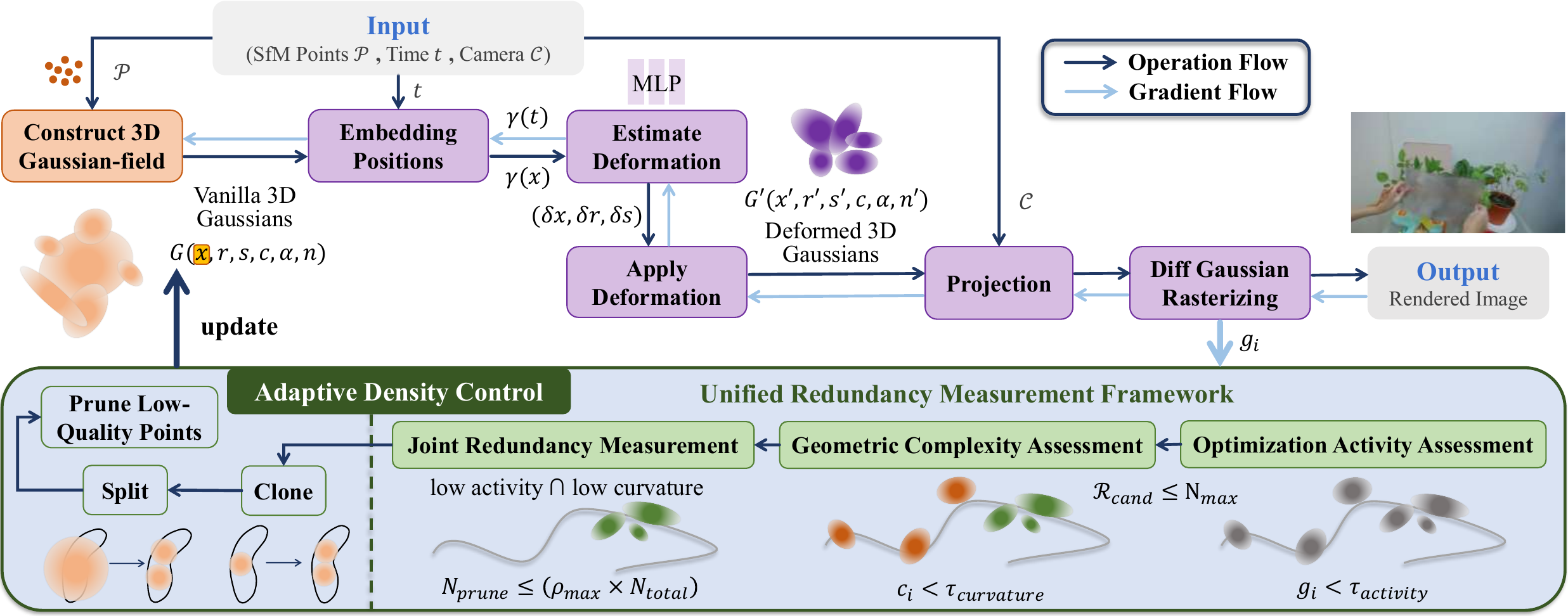} }

\caption{\textbf{The pipeline of our GARO.} The highlighted and innovative Unified Redundancy Measurement Framework consists of (1) Optimization Activity Assessment, where gray points represent low-gradient Gaussians selected as redundancy candidates using an optimization activity threshold $\tau_{activity}$, with the number of candidates limited to $N_{max}$ for computational efficiency.
(2) Geometric Complexity Assessment, where green points indicate candidate Gaussians located in flat regions with low curvature, while orange points correspond to candidate Gaussians in regions with high curvature, based on a geometric complexity threshold $\tau_{curvature}$.
(3) Joint Redundancy Measurement, where green points which have both low gradient and low curvature, are identified and pruned as redundant Gaussians, and the maximum pruning ratio $\rho_{max}$ is applied to prevent excessive removal.}
\label{pipeline}
\end{figure*}

\subsection{Overview}
As shown in Fig.~\ref{pipeline}, building upon the traditional dynamic 3D Gaussian field reconstruction pipeline, our method introduces the Unified Redundancy Measurement Framework in the adaptive density control stage. The workflow constructs a vanilla 3D Gaussian field from SfM points, encodes positions and time, uses an MLP to estimate deformation parameters, applies the deformation to obtain deformed Gaussians, and projects them onto the image plane.
Differentiable rasterization then outputs the rendered image.

Our framework identifies and prunes redundancy in three stages, including Optimization Activity Assessment, Geometric Complexity Assessment, and Joint Redundancy Measurement. Compared to traditional density control based on cloning, splitting, and pruning (left part of blue box), our approach improves efficiency by intelligently removing redundant Gaussians, achieving a better balance between reconstruction quality and computational efficiency.

\subsection{Optimization Activity Assessment}

Based on these design principles, we propose a quantitative metric for optimization activity assessment. This metric measures the average update magnitude of Gaussian parameters during training, allowing effective identification of Gaussians with minimal contribution to rendering quality.

For each Gaussian $i$, we compute the gradient magnitude of its position parameter $\mathbf{p}_i$ in view space during each training iteration and accumulate over all visible iterations to quantify long-term optimization activity:
\begin{equation}
g_i = \frac{1}{T_i} \sum_{t=1}^{T} \|\nabla_{\mathbf{p}_i} \mathcal{L}^{t}\|_2 \cdot \mathbb{I}[v_i^{t}]
,
\end{equation}
where $\mathcal{L}^{t}$ denotes the rendering loss at iteration $t$, $\mathbb{I}[\cdot]$ is the visibility indicator function, $v_i^{t}$ represents the contribution status of Gaussian $i$ to the current view at iteration $t$, which equals 1 when the projection radius exceeds 0 and 0 otherwise, and $T_i$ is the total visible count during training.

Based on this assessment strategy, we define low-gradient Gaussians as those satisfying $g_i < \tau_{activity}$, where $\tau_{activity}$ is the optimization activity threshold. These points exhibit minimal parameter changes, indicating stabilized contribution to current viewpoint rendering.
We establish these as the initial candidate set for redundant Gaussians:
\begin{equation}
\mathcal{R}_{cand} = \{i | g_i < \tau_{activity}\}
.
\end{equation}
To ensure efficient geometric complexity analysis, we limit the candidate set size to $N_{max}$ by sorting and selecting Gaussians with the lowest optimization activity:
\begin{equation}
\mathcal{R}_{\text{final}} = 
\begin{cases} 
\mathcal{R}_{\text{cand}} & \text{if } |\mathcal{R}_{\text{cand}}| \leq N_{\text{max}} \\
\text{BottomK}_{g}(\mathcal{R}_{\text{cand}}, N_{\text{max}}) & \text{otherwise}
\end{cases}
,
\end{equation}
where $\text{BottomK}$ returns the $N_{max}$ Gaussians with the lowest activity values, providing a high-quality redundant candidate set for subsequent geometric complexity analysis.

\subsection{Geometric Complexity Assessment}

After obtaining candidates with low optimization activity, we further identify truly redundant points through geometric complexity analysis. The geometric complexity metric reflects the degree of geometric variation in the local region around each Gaussian. 
The core principle is that Gaussians in flat regions are more redundant, as such areas can be represented with fewer Gaussians.

\begin{figure}[t]
\centering
\includegraphics[width=0.95\columnwidth]{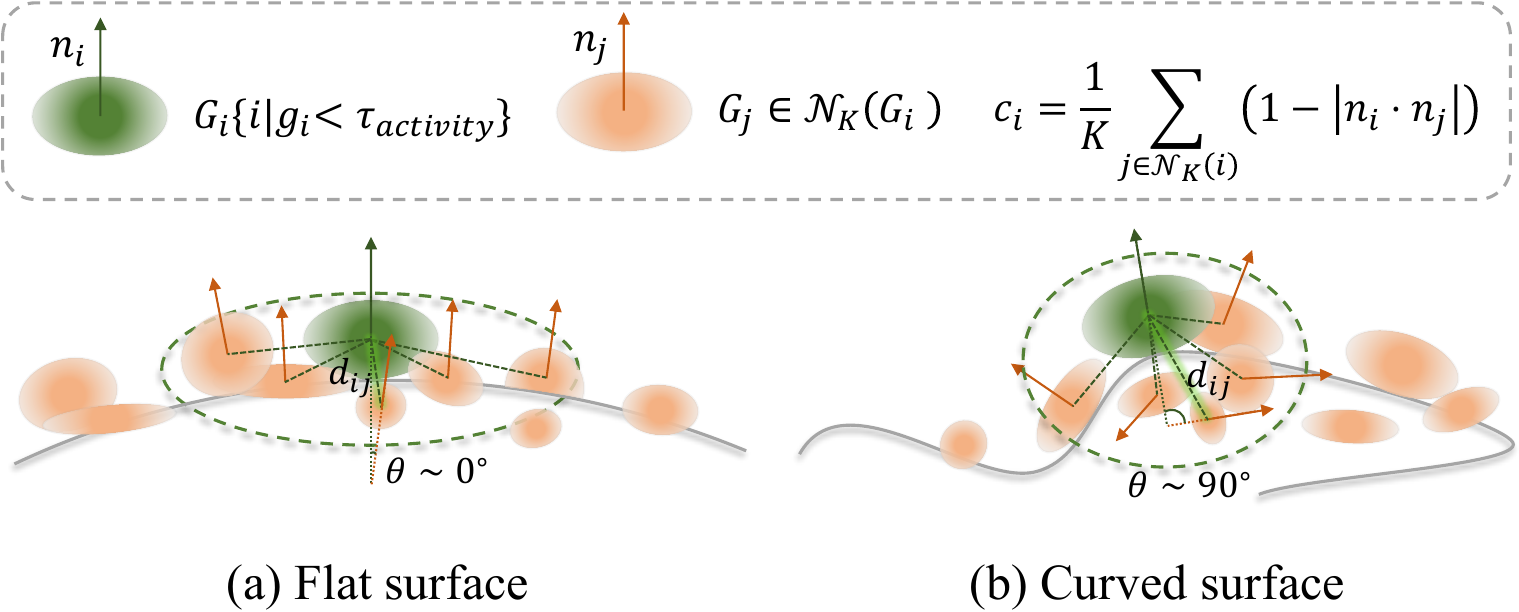} 
\caption{\textbf{Curvature computation process for Gaussian primitives on flat and curved surfaces.} The green Gaussian represents the target point with low optimization activity requiring curvature estimation. We compute normal vector differences with K neighboring points and average them to estimate local curvature. (a) On flat surfaces, pairwise normal differences are minimal, resulting in low curvature. (b) On curved surfaces, pairwise normal differences are significant, yielding high curvature values.}
\label{method}
\end{figure}

To support geometric complexity analysis, we establish a normal vector for each Gaussian. Based on the geometric properties of Gaussian ellipsoids, we use the shortest axis direction as the initial normal vector estimation:
\begin{equation}
    \mathbf{n}_i = R_i[:, k], \quad k = \arg\min ([s_1, s_2, s_3])
,    
\label{eq:最短轴近似}
\end{equation}
where $R_i$ is the rotation matrix and $s_i$ are the scaling parameters. During dynamic deformation, to maintain consistency between the normal vectors and the local geometry, we employ a coordinate system transformation-based update strategy:
\begin{equation}
    \mathbf{n}_i' = \mathbf{V}_i\mathbf{U}_i^T \mathbf{n}_i
,
\end{equation}
where $\mathbf{U}_i$ and $\mathbf{V}_i$ are the local orthogonal bases before and after deformation, constructed from ellipsoid scaling and rotation parameters.
Inspired by point cloud surface curvature estimation theory \cite{ESPS}, we adopt a normal variation-based curvature metric to quantify local geometric complexity:
\begin{equation}
    c_i = \frac{1}{K} \sum_{j \in \mathcal{N}_K(i)} \left(1 - |\mathbf{n}_i \cdot \mathbf{n}_j|\right)
,
\end{equation}
where $\mathcal{N}_K(i)$ represents the $K$ nearest neighbors ($K=10$) of Gaussian $i$, and $\mathbf{n}_i$, $\mathbf{n}_j$ are the corresponding unit normal vectors. According to differential geometry theory \cite{curves}, local surface curvature can be approximated through the variation degree of adjacent normal vectors. Specifically, we compute the angular differences between K pairs of normal vectors and take their average to obtain the curvature estimation. As illustrated in Fig.~\ref{method}, when adjacent normal vectors are highly consistent, the curvature value approaches 0, indicating low geometric complexity. Conversely, when adjacent normal vectors exhibit significant angular differences, the curvature value approaches 1, indicating high geometric complexity.

According to this assessment, we define low geometric complexity Gaussians as those satisfying $c_i < \tau_{curvature}$, where $\tau_{curvature}$ is the geometric complexity threshold. These Gaussians are located in regions with gentle local geometric variation, where the represented geometric information exhibits strong spatial continuity and predictability. 

\subsection{Joint Redundancy Measurement}

Through independent assessments of optimization activity and geometric complexity, we construct a joint redundancy measurement framework that ensures reliable pruning decisions through dual-condition constraints: 
\begin{equation}
    r_i = \mathbb{I}[g_i < \tau_{activity}] \cdot \mathbb{I}[c_i < \tau_{curvature}]
,
\end{equation}
where $\mathbb{I}[\cdot]$ denotes the indicator function. When $r_i = 1$, Gaussian $i$ simultaneously satisfies both low optimization activity and low geometric complexity conditions, qualifying as a redundant candidate.

To prevent excessive pruning that could compromise scene representation continuity, we introduce an adaptive pruning ratio control mechanism. The number of Gaussians actually pruned is constrained by both redundancy identification and scene-adaptive ratio limits:
\begin{equation}
    N_{prune} = \min\left(\sum_i r_i, \lfloor \rho_{max} \times N_{total} \rfloor\right)
,
\end{equation}
where $\rho_{max}$ is the maximum pruning ratio and $N_{total}$ is the current total number of Gaussians. Among identified redundant candidates, we prioritize pruning those with the lowest curvature values to ensure maximum redundancy removal: 
\begin{equation}
    \mathcal{P}_{final} = \text{BottomK}_{c}(\{i | r_i = 1\}, N_{prune})
.
\end{equation}
Building on this unified redundancy measurement, our approach employs dual-condition constraints and hierarchical filtering to ensure accurate and safe pruning of redundant Gaussians. By first screening low-activity candidates and then computing curvature only for selected points, we significantly reduce computational overhead while preserving geometrically important structures. The synergy of optimization activity and geometric complexity enables efficient and reliable reconstruction of complex dynamic scenes.

%% file: tex/Experiments.tex
\section{Experiments}
\begin{figure*}[htbp]
    \centering 
    \includegraphics[width=0.98\textwidth]{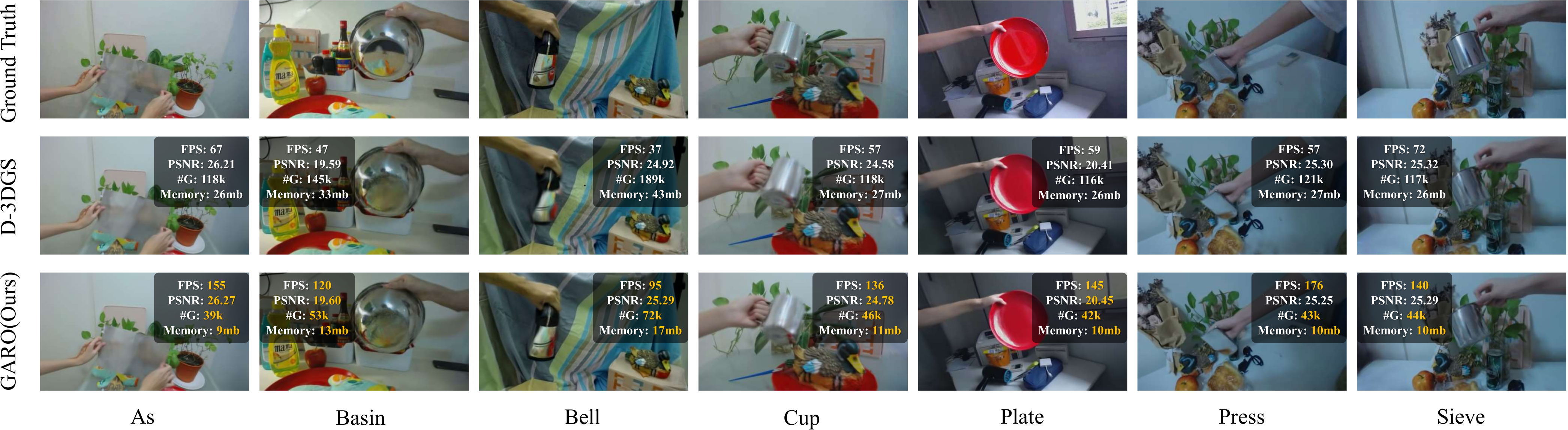}
    \caption{\textbf{Qualitative comparison on the NeRF-DS dataset between ground truth (top), D-3DGS (middle), and our GARO (bottom).} All improved metrics are highlighted in orange. In all scenes, our method significantly reduces the number of Gaussians, achieving substantial improvements in FPS and memory usage while maintaining comparable or even better rendering quality.} 
    \label{qualitative_nerfds}
\end{figure*}

\subsection{Experimental Setup}

\subsubsection{Dataset and metrics}

To validate our method, we conduct experiments on two representative dynamic scene datasets. D-NeRF \cite{D-NeRF} includes eight synthetic dynamic scenes with varying camera poses at each timestep, while NeRF-DS \cite{NeRF-DS} contains seven real-world dynamic scenes with complex reflections and specular objects. We use standard image quality metrics including Peak Signal-to-Noise Ratio (PSNR), Structural Similarity (SSIM) \cite{SSIM}, and Learned Perceptual Image Patch Similarity (LPIPS) \cite{LPIPS} to assess rendering fidelity. We also report frames per second (FPS) for computational efficiency, as well as memory consumption (Memory) and the number of Gaussians (\#G).

\subsubsection{Implementation}
Our framework is implemented in PyTorch \cite{PyTorch} and trained for 20k iterations on NeRF-DS and 40k iterations on D-NeRF. 
The training and testing splits, as well as image resolutions, are consistent with the original datasets. All experiments are conducted on a single NVIDIA RTX 3090 GPU. Key hyperparameters include the optimization activity threshold $\tau_{activity}$ of 0.00005, the maximum candidate point limit $N_{max}$ of 10k, the geometric complexity threshold $\tau_{curvature}$ of 0.5, and the maximum pruning ratio $\rho_{max}$ of 0.02. Other hyperparameters follow default settings or those used in prior work.

\subsection{Quantitative Comparisons}
\label{quantitative}

\subsubsection{D-NeRF dataset}
\label{quantitative_dnerf}
We compare our method with several baselines that are most relevant to our work. As shown in Tab.~\ref{table:dnerf}, these methods are divided into two blocks. The upper block includes neural implicit models such as D-NeRF \cite{D-NeRF}, TiNeuVox-B \cite{TiNeuVox-B}, Tensor4D \cite{Tensor4D}, KPlanes \cite{K-Planes}, HexPlane-Slim \cite{HexPlane}, and Ti-DNeRF \cite{ti-nerf}, while the lower block contains explicit Gaussian-based approaches including 4DGS \cite{4DGS_wu}, D-3DGS \cite{D-3DGS}, and SP-GS+NG \cite{SP-GS}.
For D-3DGS, we report results from our own reproduction using the official implementation, as the original paper does not provide memory metrics. Minor differences from the original results may occur due to environment variations, but this ensures fair and consistent comparison with our method. SP-GS framework includes both SP-GS and SP-GS+NG models. SP-GS+NG integrates a non-rigid deformation network for dynamic scenes, similar to our method which predicts offsets for position, rotation, and scaling via a deformation MLP. Therefore, we compare our results with SP-GS+NG for fairness.

\begin{table}[t]
    \centering
    \begin{tabular}{c|*{4}{c}}
        \toprule
        Methods & PSNR$\uparrow$ & SSIM$\uparrow$ & LPIPS$\downarrow$ & FPS$\uparrow$ \\ \midrule
        D-NeRF & 31.14 &  0.9761 & 0.0464 & $<1$   \\
        TiNeuVox-B & 32.74  & 0.9715 & 0.0495 & $\sim$ 1.5 \\
        Tensor4D & 27.44 &  0.9421  &  0.0569 & -\\
        KPlanes & 31.41 & 0.9699  & 0.0470 & $\sim$ 0.12 \\
        HexPlane-Slim & 32.97 & 0.9750 & 0.0346 & 4 \\
        Ti-DNeRF & 32.69 & 0.9746 & 0.3580 & -  \\
        \midrule
        4DGS & 35.31  & 0.9841  & \cellcolor{third}{0.0148} & 144 \\
        D-3DGS & \cellcolor{best}{40.22} & \cellcolor{best}{0.9913} & \cellcolor{best}{0.0127} & \cellcolor{second}147 \\
        SP-GS+NG & \cellcolor{third}{38.28}  & \cellcolor{third}{0.9877}  & 0.0152 & \cellcolor{third}119  \\
        GARO(ours) & \cellcolor{second}39.76  & \cellcolor{second}0.9908  & \cellcolor{second}0.0143 & \cellcolor{best}{295} \\
        \bottomrule
    \end{tabular}
\caption{\textbf{Quantitative comparison on D-NeRF dataset.} We report the average PSNR, SSIM, and LPIPS of baseline methods. Results for these methods are cited from SP-GS \cite{SP-GS} except D-3DGS which we reproduced using the official implementation. We color each cell as \colorbox{best}{best}, \colorbox{second}{second best}, and \colorbox{third}{third best}. ‘-’ denotes that the metric is not reported in their
works. Lego is excluded.}
\label{table:dnerf}
\end{table}

The quantitative results on the D-NeRF dataset are summarized in Tab.~\ref{table:dnerf}. We report the average PSNR, SSIM, LPIPS, and FPS across all scenes, excluding the Lego scene due to a significant distribution mismatch between its training and test sets that leads to abnormal reconstruction. 
Our method achieves a significant improvement in rendering speed, reaching 295 FPS, which is $2\times$ that of D-3DGS and substantially higher than all other baselines, while maintaining PSNR reduction within 0.5 dB and SSIM and LPIPS values comparable to D-3DGS.

\begin{figure}[ht]
\centering
\includegraphics[width=0.95\columnwidth]{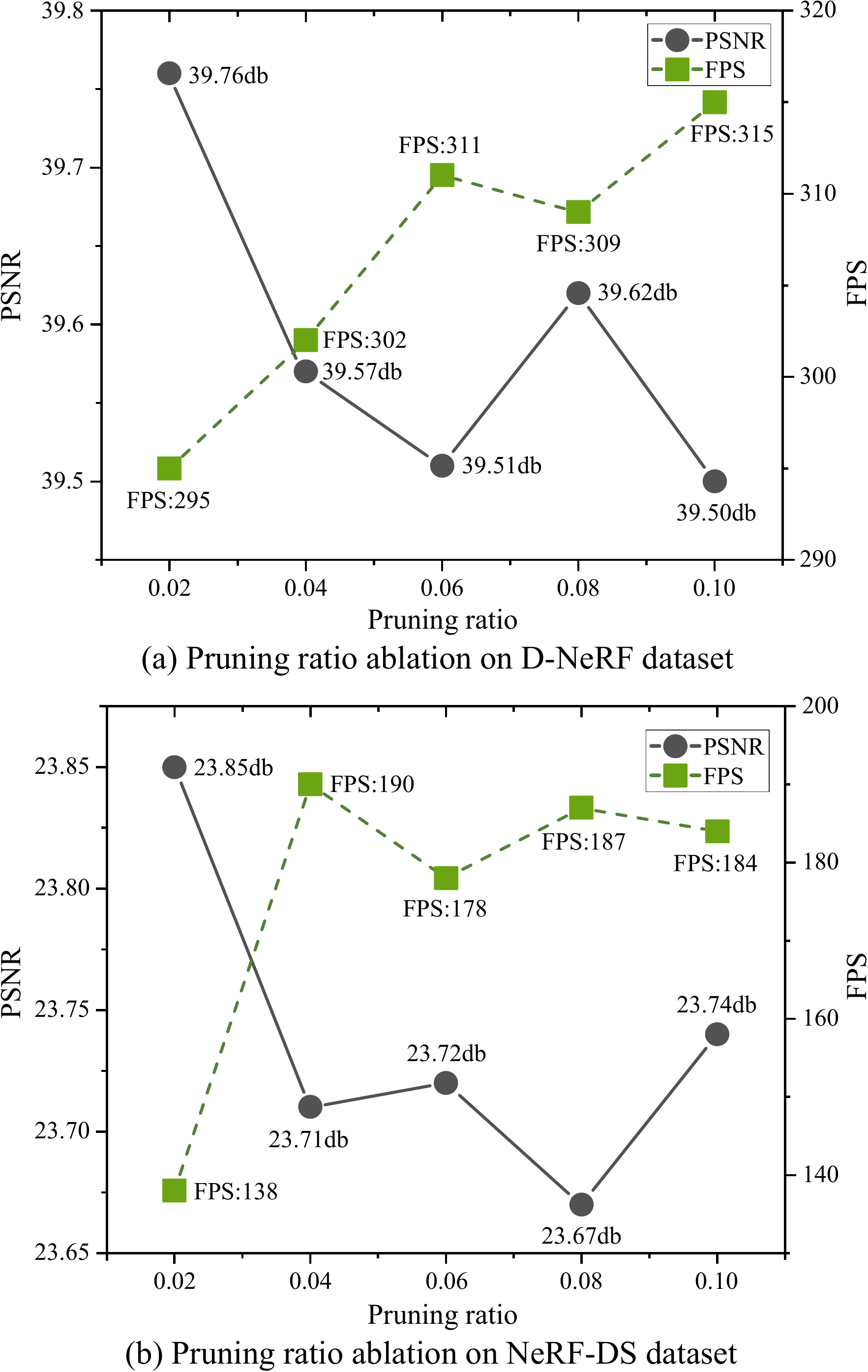} 
\caption{\textbf{Ablation study of different pruning ratios on rendering performance.} GARO demonstrates consistent quality-speed trade-offs on both D-NeRF and NeRF-DS datasets. Overall, the pruning ratio is positively correlated with FPS and negatively correlated with PSNR, but the PSNR variation across all pruning ratios is controlled within 0.3 dB, validating the robustness of the pruning strategy.}
\label{ablation_prune_ratio}
\end{figure}

\subsubsection{NeRF-DS dataset}
\label{quantitative_nerfds}

We further evaluate GARO on the NeRF-DS \cite{NeRF-DS} dataset, comparing against baseline methods. We also divide these methods into two blocks in Tab.~\ref{table:nerfds}, including TiNeuVox \cite{TiNeuVox-B}, HyperNeRF \cite{HyperNeRF}, NeRF-DS, D-3DGS \cite{D-3DGS}, and SP-GS+NG \cite{SP-GS}. 
We report average PSNR, SSIM, LPIPS, and FPS across all scenes. Our method achieves a remarkable improvement in rendering speed, reaching 138 FPS, which is $2\times$ that of D-3DGS and substantially higher than SP-GS+NG. Notably, our approach also achieves the best PSNR, SSIM, and LPIPS among all methods. This superior performance demonstrates that our GARO not only accelerates dynamic scene rendering but also enhances visual fidelity.

\begin{table}[t]
    \centering
    \begin{tabular}{c|*{4}{c}}
        \toprule
        Methods & PSNR$\uparrow$ & SSIM$\uparrow$ & LPIPS$\downarrow$ & FPS$\uparrow$ \\ \midrule
        TiNeuVox & 21.61  & 0.8241 & 0.3195 & - \\
        HyperNeRF & 23.45 & 0.8488  & 0.2002 & - \\
        NeRF-DS & \cellcolor{third}23.60 & \cellcolor{third}0.8494  & \cellcolor{third}0.1816 & - \\
        \midrule
        D-3DGS & \cellcolor{second}{23.76} & \cellcolor{second}{0.8504} & \cellcolor{second}{0.1800} & \cellcolor{third}57 \\
        SP-GS+NG & 23.33  & 0.8362  & 0.2084 & \cellcolor{second}{66} \\
        GARO(ours) & \cellcolor{best}{23.85}  & \cellcolor{best}{0.8522}   & \cellcolor{best}0.1799 & \cellcolor{best}138  \\
        \bottomrule
    \end{tabular}
\caption{\textbf{Quantitative comparison on NeRF-DS dataset.} We report the average PSNR, SSIM, and LPIPS of baseline methods. Results for these methods are cited from SP-GS \cite{SP-GS} except D-3DGS which we reproduced using the official implementation. We color each cell as \colorbox{best}{best}, \colorbox{second}{second best}, and \colorbox{third}{third best}.}
\label{table:nerfds}
\end{table}

\subsection{Qualitative Comparisons}
\label{quanlitative}
Qualitative comparisons on the NeRF-DS dataset are shown in Fig.~\ref{qualitative_nerfds}. As illustrated, our GARO method consistently achieves significant improvements in rendering speed and memory efficiency across all scenes, with the most notable gains observed in As, Plate, and Cup.
Importantly, GARO maintains comparable or even better PSNR in these scenes.

\subsection{Ablation Studies}

\subsubsection{Effect of Pruning Ratio on Quality and Efficiency}
To evaluate the impact of different maximum pruning ratios $\rho_{max}$ on rendering performance, we conduct ablation experiments with five pruning settings. As shown in Fig.~\ref{ablation_prune_ratio}, results on both datasets indicate that as the $\rho_{max}$ increases, PSNR gradually decreases while FPS consistently improves, demonstrating a stable quality-speed trade-off. Notably, in real-world scenarios, the PSNR drop relative to D-3DGS is strictly controlled within 0.5 dB across all configurations, validating the precision and robustness of our method. We select 0.02 as the final pruning ratio to ensure the best reconstruction quality while still achieving a significant speedup.

\subsubsection{Ablation on Optimization Activity and Geometric Complexity Assessment}

\begin{table}[t]
    \centering
    \setlength{\tabcolsep}{0.8mm}
\begin{tabular}{c|ccc|ccc}
      \toprule
      \multirow{1}[3]{*}{Methods}& \multicolumn{3}{c|}{D-NeRF} & \multicolumn{3}{c}{NeRF-DS}\\
            & \multicolumn{1}{l}{PSNR$\uparrow$} & \multicolumn{1}{l}{\#G(k)$\downarrow$} & \multicolumn{1}{l|}{FPS$\uparrow$} & \multicolumn{1}{l}{PSNR$\uparrow$} & \multicolumn{1}{l}{\#G(k)$\downarrow$} & \multicolumn{1}{l}{FPS$\uparrow$}\\
      \midrule
      w/o GCA& 39.68 & 19 & 281 & 23.68 & 48 & 140\\
      w/o OAA& 39.59 & 20 & 270 & 23.77 & 57 & 124\\
      \midrule
      full model & 39.76 & 18 & 295 & 23.85 & 48 & 138\\
      \bottomrule
      \end{tabular}
\caption{\textbf{Ablation study of redundancy identification strategies.} OAA denotes Optimization Activity Assessment, which uses only gradient-based pruning. GCA denotes Geometric Complexity Assessment, which uses only curvature-based pruning. \#G denotes the number of Gaussians.}
\label{table:ablation}
\end{table}

To investigate the effectiveness and complementarity of different redundancy identification strategies, we conduct ablation experiments on both the D-NeRF and NeRF-DS datasets. All strategies use the same maximum pruning ratio $\rho_{max}$ of 0.02 to ensure a fair comparison of redundancy identification accuracy. Under this constraint, the results in Tab.~\ref{table:ablation} show that the full model, which combines both OAA and GCA strategies, consistently achieves the highest reconstruction quality and the lowest number of Gaussians. This demonstrates that the joint redundancy measurement framework outperforms either single strategy in accurately identifying redundant Gaussians, leading to a more compact and expressive model.

%% file: tex/Conclusion.tex
\section{Conclusion}
In conclusion, we tackle the efficiency challenges in dynamic novel view synthesis by introducing geometry-aware redundancy optimization.
Our core innovation is a unified redundancy measurement framework integrated into the adaptive density control stage of the dynamic reconstruction pipeline, which systematically prunes unnecessary Gaussians through joint analysis of optimization activity and geometric structure.
Experiments on both synthetic and real-world datasets show substantial gains in rendering speed and memory usage while maintaining reconstruction quality. 

\section*{ACKNOWLEDGMENT}
The authors gratefully acknowledge financial support. Kaixing Zhao is supported by the National Natural Science Foundation of China under Grant No. 62407035.

%% file: GARO.bbl
\begin{thebibliography}{10}
\providecommand{\url}[1]{#1}
\csname url@rmstyle\endcsname
\providecommand{\newblock}{\relax}
\providecommand{\bibinfo}[2]{#2}
\providecommand\BIBentrySTDinterwordspacing{\spaceskip=0pt\relax}
\providecommand\BIBentryALTinterwordstretchfactor{4}
\providecommand\BIBentryALTinterwordspacing{\spaceskip=\fontdimen2\font plus
\BIBentryALTinterwordstretchfactor\fontdimen3\font minus \fontdimen4\font\relax}
\providecommand\BIBforeignlanguage[2]{{%
\expandafter\ifx\csname l@#1\endcsname\relax
\typeout{** WARNING: IEEEtran.bst: No hyphenation pattern has been}%
\typeout{** loaded for the language `#1'. Using the pattern for}%
\typeout{** the default language instead.}%
\else
\language=\csname l@#1\endcsname
\fi
#2}}

\bibitem{LidaRF}
S.~Sun, B.~Zhuang, Z.~Jiang, B.~Liu, X.~Xie, and M.~Chandraker, ``Lidarf: Delving into lidar for neural radiance field on street scenes,'' in \emph{Proceedings of the IEEE/CVF Conference on Computer Vision and Pattern Recognition}, 2024, pp. 19\,563--19\,572.

\bibitem{FastScene}
Y.~Ma, D.~Zhan, and Z.~Jin, ``{FastScene}: {Text-Driven} {Fast} {3D} {Indoor} {Scene} {Generation} via {Panoramic} {Gaussian} {Splatting},'' in \emph{Proceedings of the 33rd International Joint Conference on Artificial Intelligence}, ser. IJCAI '24, Jeju, South Korea, 2024, pp. 1173--1181.

\bibitem{Palm}
A.~Zhou, M.~J. Kim, L.~Wang, P.~Florence, and C.~Finn, ``{NeRF} in the palm of your hand: {Corrective} augmentation for robotics via novel-view synthesis,'' in \emph{Proceedings of the IEEE/CVF Conference on Computer Vision and Pattern Recognition}, 2023, pp. 17\,907--17\,917.

\bibitem{Nerf}
B.~Mildenhall, P.~P. Srinivasan, M.~Tancik, J.~T. Barron, R.~Ramamoorthi, and R.~Ng, ``Nerf: Representing scenes as neural radiance fields for view synthesis,'' in \emph{Proceedings of the European Conference on Computer Vision}, vol. 12346, 2020, pp. 405--421.

\bibitem{3dgs}
B.~Kerbl, G.~Kopanas, T.~Leimk{\"{u}}hler, and G.~Drettakis, ``3d gaussian splatting for real-time radiance field rendering,'' \emph{{ACM} Trans. Graph.}, vol.~42, no.~4, pp. 139:1--139:14, 2023.

\bibitem{Dynamic}
J.~Luiten, G.~Kopanas, B.~Leibe, and D.~Ramanan, ``Dynamic 3d gaussians: Tracking by persistent dynamic view synthesis,'' in \emph{Proceedings of the International Conference on 3D Vision}, 2024, pp. 800--809.

\bibitem{Mip-Splatting}
Z.~Yu, A.~Chen, B.~Huang, T.~Sattler, and A.~Geiger, ``Mip-splatting: Alias-free 3d gaussian splatting,'' in \emph{Proceedings of the IEEE/CVF Conference on Computer Vision and Pattern Recognition}, 2024, pp. 19\,447--19\,456.

\bibitem{IRGS}
C.~Gu, X.~Wei, Z.~Zeng, Y.~Yao, and L.~Zhang, ``{IRGS:} inter-reflective gaussian splatting with 2d gaussian ray tracing,'' in \emph{Proceedings of the Computer Vision and Pattern Recognition Conference}, 2025, pp. 10\,943--10\,952.

\bibitem{LightGaussian}
Z.~Fan, K.~Wang, K.~Wen, Z.~Zhu, D.~Xu, Z.~Wang, \emph{et~al.}, ``Lightgaussian: Unbounded 3d gaussian compression with 15x reduction and 200+ fps,'' \emph{Advances in neural information processing systems}, vol.~37, pp. 140\,138--140\,158, 2024.

\bibitem{CompGS}
K.~L. Navaneet, K.~P. Meibodi, S.~A. Koohpayegani, and H.~Pirsiavash, ``Compgs: Smaller and faster gaussian splatting with vector quantization,'' in \emph{European Conference on Computer Vision}, vol. 15090, 2024, pp. 330--349.

\bibitem{Compressed}
S.~Niedermayr, J.~Stumpfegger, and R.~Westermann, ``Compressed 3d gaussian splatting for accelerated novel view synthesis,'' in \emph{Proceedings of the IEEE/CVF Conference on Computer Vision and Pattern Recognition}, 2024, pp. 10\,349--10\,358.

\bibitem{Reducing}
P.~Papantonakis, G.~Kopanas, B.~Kerbl, A.~Lanvin, and G.~Drettakis, ``Reducing the memory footprint of 3d gaussian splatting,'' \emph{Proc. {ACM} Comput. Graph. Interact. Tech.}, vol.~7, no.~1, pp. 16:1--16:17, 2024.

\bibitem{HAC}
Y.~Chen, Q.~Wu, W.~Lin, M.~Harandi, and J.~Cai, ``{HAC:} hash-grid assisted context for 3d gaussian splatting compression,'' in \emph{European Conference on Computer Vision}, vol. 15065, 2024, pp. 422--438.

\bibitem{PUP}
A.~Hanson, A.~Tu, V.~Singla, M.~Jayawardhana, M.~Zwicker, and T.~Goldstein, ``{PUP} 3d-gs: Principled uncertainty pruning for 3d gaussian splatting,'' in \emph{Proceedings of the Computer Vision and Pattern Recognition Conference}, 2025, pp. 5949--5958.

\bibitem{Compact}
J.~C. Lee, D.~Rho, X.~Sun, J.~H. Ko, and E.~Park, ``Compact 3d gaussian representation for radiance field,'' in \emph{Proceedings of the IEEE/CVF Conference on Computer Vision and Pattern Recognition}, 2024, pp. 21\,719--21\,728.

\bibitem{LP-3DGS}
Z.~Zhang, T.~Song, Y.~Lee, L.~Yang, C.~Peng, R.~Chellappa, and D.~Fan, ``{LP-3DGS:} learning to prune 3d gaussian splatting,'' in \emph{Advances in Neural Information Processing Systems}, vol.~37, 2024, pp. 122\,434--122\,457.

\bibitem{Radsplat}
M.~Niemeyer, F.~Manhardt, M.-J. Rakotosaona, M.~Oechsle, D.~Duckworth, R.~Gosula, K.~Tateno, J.~Bates, D.~Kaeser, and F.~Tombari, ``Radsplat: Radiance field-informed gaussian splatting for robust real-time rendering with 900+ fps,'' \emph{arXiv preprint arXiv:2403.13806}, 2024.

\bibitem{Mini-splatting}
G.~Fang and B.~Wang, ``Mini-splatting: Representing scenes with a constrained number of gaussians,'' in \emph{European Conference on Computer Vision}.\hskip 1em plus 0.5em minus 0.4em\relax Springer, 2024, pp. 165--181.

\bibitem{Taming}
S.~S. Mallick, R.~Goel, B.~Kerbl, M.~Steinberger, F.~V. Carrasco, and F.~De~La~Torre, ``Taming 3dgs: High-quality radiance fields with limited resources,'' in \emph{SIGGRAPH Asia 2024 Conference Papers}, 2024, pp. 1--11.

\bibitem{Scaffold-GS}
T.~Lu, M.~Yu, L.~Xu, Y.~Xiangli, L.~Wang, D.~Lin, and B.~Dai, ``Scaffold-gs: Structured 3d gaussians for view-adaptive rendering,'' in \emph{Proceedings of the IEEE/CVF Conference on Computer Vision and Pattern Recognition}, 2024, pp. 20\,654--20\,664.

\bibitem{DiffGS}
J.~Zhou, W.~Zhang, and Y.~Liu, ``Diffgs: Functional gaussian splatting diffusion,'' in \emph{Advances in Neural Information Processing Systems}, 2024.

\bibitem{Compact_dy}
K.~Katsumata, D.~M. Vo, and H.~Nakayama, ``A compact dynamic 3d gaussian representation for real-time dynamic view synthesis,'' in \emph{European Conference on Computer Vision}, ser. Lecture Notes in Computer Science, vol. 15144, 2024, pp. 394--412.

\bibitem{MaskGaussian}
Y.~Liu, Z.~Zhong, Y.~Zhan, S.~Xu, and X.~Sun, ``Maskgaussian: Adaptive 3d gaussian representation from probabilistic masks,'' in \emph{Proceedings of the Computer Vision and Pattern Recognition Conference}, 2025, pp. 681--690.

\bibitem{EDGS}
H.~Kong, X.~Yang, and X.~Wang, ``Efficient gaussian splatting for monocular dynamic scene rendering via sparse time-variant attribute modeling,'' in \emph{Proceedings of the AAAI Conference on Artificial Intelligence}, 2025, pp. 4374--4382.

\bibitem{SP-GS}
D.~Wan, R.~Lu, and G.~Zeng, ``Superpoint gaussian splatting for real-time high-fidelity dynamic scene reconstruction,'' in \emph{Proceedings of the Forty-first International Conference on Machine Learning ({ICML} 2024)}, 2024.

\bibitem{RTGuIDE}
Y.~Tao, D.~Ong, V.~Murali, I.~Spasojevic, P.~Chaudhari, and V.~Kumar, ``Rt-guide: Real-time gaussian splatting for information-driven exploration,'' \emph{IEEE Robotics and Automation Letters}, vol.~10, no.~11, pp. 11\,594--11\,601, 2025.

\bibitem{sfm}
J.~L. Sch{\"{o}}nberger and J.~Frahm, ``Structure-from-motion revisited,'' in \emph{Proceedings of the IEEE conference on computer vision and pattern recognition}, 2016, pp. 4104--4113.

\bibitem{mvs}
J.~L. Sch{\"o}nberger, E.~Zheng, J.-M. Frahm, and M.~Pollefeys, ``Pixelwise view selection for unstructured multi-view stereo,'' in \emph{European conference on computer vision}.\hskip 1em plus 0.5em minus 0.4em\relax Springer, 2016, pp. 501--518.

\bibitem{D-NeRF}
A.~Pumarola, E.~Corona, G.~Pons{-}Moll, and F.~Moreno{-}Noguer, ``D-nerf: Neural radiance fields for dynamic scenes,'' in \emph{Proceedings of the IEEE/CVF Conference on Computer Vision and Pattern Recognition}, 2021, pp. 10\,318--10\,327.

\bibitem{NeRF-DS}
Z.~Yan, C.~Li, and G.~H. Lee, ``Nerf-ds: Neural radiance fields for dynamic specular objects,'' in \emph{Proceedings of the IEEE/CVF Conference on Computer Vision and Pattern Recognition}, 2023, pp. 8285--8295.

\bibitem{Nerfies}
K.~Park, U.~Sinha, J.~T. Barron, S.~Bouaziz, D.~B. Goldman, S.~M. Seitz, and R.~Martin{-}Brualla, ``Nerfies: Deformable neural radiance fields,'' in \emph{Proceedings of the IEEE/CVF International Conference on Computer Vision}, 2021, pp. 5845--5854.

\bibitem{HyperNeRF}
K.~Park, U.~Sinha, P.~Hedman, J.~T. Barron, S.~Bouaziz, D.~B. Goldman, R.~Martin{-}Brualla, and S.~M. Seitz, ``Hypernerf: a higher-dimensional representation for topologically varying neural radiance fields,'' \emph{{ACM} Trans. Graph.}, vol.~40, no.~6, pp. 238:1--238:12, 2021.

\bibitem{ti-nerf}
S.~Park, M.~Son, S.~Jang, Y.~C. Ahn, J.~Kim, and N.~Kang, ``Temporal interpolation is all you need for dynamic neural radiance fields,'' in \emph{Proceedings of the IEEE/CVF conference on computer vision and pattern recognition}, 2023, pp. 4212--4221.

\bibitem{TiNeuVox-B}
J.~Fang, T.~Yi, X.~Wang, L.~Xie, X.~Zhang, W.~Liu, M.~Nie{\ss}ner, and Q.~Tian, ``Fast dynamic radiance fields with time-aware neural voxels,'' in \emph{Proceedings of SIGGRAPH Asia 2022 Conference Papers}, 2022, pp. 11:1--11:9.

\bibitem{Tensor4D}
R.~Shao, Z.~Zheng, H.~Tu, B.~Liu, H.~Zhang, and Y.~Liu, ``Tensor4d: Efficient neural 4d decomposition for high-fidelity dynamic reconstruction and rendering,'' in \emph{Proceedings of the IEEE/CVF Conference on Computer Vision and Pattern Recognition}, 2023, pp. 16\,632--16\,642.

\bibitem{K-Planes}
S.~Fridovich{-}Keil, G.~Meanti, F.~R. Warburg, B.~Recht, and A.~Kanazawa, ``K-planes: Explicit radiance fields in space, time, and appearance,'' in \emph{Proceedings of the IEEE/CVF Conference on Computer Vision and Pattern Recognition}, 2023, pp. 12\,479--12\,488.

\bibitem{HexPlane}
A.~Cao and J.~Johnson, ``Hexplane: {A} fast representation for dynamic scenes,'' in \emph{Proceedings of the IEEE/CVF Conference on Computer Vision and Pattern Recognition}, 2023, pp. 130--141.

\bibitem{Spacetime}
Z.~Li, Z.~Chen, Z.~Li, and Y.~Xu, ``Spacetime gaussian feature splatting for real-time dynamic view synthesis,'' in \emph{Proceedings of the IEEE/CVF Conference on Computer Vision and Pattern Recognition}, 2024, pp. 8508--8520.

\bibitem{D-3DGS}
Z.~Yang, X.~Gao, W.~Zhou, S.~Jiao, Y.~Zhang, and X.~Jin, ``Deformable 3d gaussians for high-fidelity monocular dynamic scene reconstruction,'' in \emph{Proceedings of the IEEE/CVF Conference on Computer Vision and Pattern Recognition}, 2024, pp. 20\,331--20\,341.

\bibitem{4DGS_Yang}
Z.~Yang, H.~Yang, Z.~Pan, and L.~Zhang, ``Real-time photorealistic dynamic scene representation and rendering with 4d gaussian splatting,'' in \emph{Proceedings of the Twelfth International Conference on Learning Representations}, 2024.

\bibitem{4DGS_wu}
G.~Wu, T.~Yi, J.~Fang, L.~Xie, X.~Zhang, W.~Wei, W.~Liu, Q.~Tian, and X.~Wang, ``4d gaussian splatting for real-time dynamic scene rendering,'' in \emph{Proceedings of the IEEE/CVF Conference on Computer Vision and Pattern Recognition}, 2024, pp. 20\,310--20\,320.

\bibitem{SC-GS}
Y.~Huang, Y.~Sun, Z.~Yang, X.~Lyu, Y.~Cao, and X.~Qi, ``{SC-GS:} sparse-controlled gaussian splatting for editable dynamic scenes,'' in \emph{Proceedings of the IEEE/CVF conference on computer vision and pattern recognition}, 2024, pp. 4220--4230.

\bibitem{FDS-GS}
Z.~Zeng, Y.~Wang, L.~Ju, and T.~Guan, ``Frequency-aware density control via reparameterization for high-quality rendering of 3d gaussian splatting,'' in \emph{Proceedings of the AAAI Conference on Artificial Intelligence}, 2025, pp. 9833--9841.

\bibitem{ESPS}
M.~Pauly, M.~H. Gross, and L.~Kobbelt, ``Efficient simplification of point-sampled surfaces,'' in \emph{IEEE Visualization, 2002. VIS 2002.}, 2002, pp. 163--170.

\bibitem{curves}
M.~P. do~Carmo, \emph{Differential geometry of curves and surfaces}.\hskip 1em plus 0.5em minus 0.4em\relax Prentice Hall, 1976.

\bibitem{SSIM}
Z.~Wang, A.~C. Bovik, H.~R. Sheikh, and E.~P. Simoncelli, ``Image quality assessment: from error visibility to structural similarity,'' \emph{{IEEE} Trans. Image Process.}, vol.~13, no.~4, pp. 600--612, 2004.

\bibitem{LPIPS}
R.~Zhang, P.~Isola, A.~A. Efros, E.~Shechtman, and O.~Wang, ``The unreasonable effectiveness of deep features as a perceptual metric,'' in \emph{Proceedings of the IEEE/CVF Conference on Computer Vision and Pattern Recognition}, 2018, pp. 586--595.

\bibitem{PyTorch}
A.~Paszke, S.~Gross, F.~Massa, A.~Lerer, J.~Bradbury, G.~Chanan, T.~Killeen, Z.~Lin, N.~Gimelshein, L.~Antiga, \emph{et~al.}, ``{PyTorch}: {An} {Imperative} {Style}, {High-Performance} {Deep} {Learning} {Library},'' in \emph{Advances in Neural Information Processing Systems}, ser. NeurIPS '19, vol.~32, 2019, pp. 8026--8037.

\end{thebibliography}
